\documentclass[10pt,twocolumn,letterpaper]{article}

\usepackage[pagenumbers]{cvpr} 

\definecolor{cvprblue}{rgb}{0.21,0.49,0.74}
\usepackage[pagebackref,breaklinks,colorlinks,allcolors=cvprblue]{hyperref}
\usepackage{array} 
\newcolumntype{L}[1]{>{\raggedright\arraybackslash}p{#1}} 
\newcolumntype{C}[1]{>{\centering\arraybackslash}p{#1}}
\newcolumntype{R}[1]{>{\raggedleft\arraybackslash}p{#1}}

\usepackage{amsmath}
\usepackage{multirow}
\usepackage{newfloat}
\usepackage{listings}
\usepackage{booktabs}
\usepackage{colortbl}
\definecolor{Gray}{gray}{0.9}

\usepackage{adjustbox}

\def\confName{CVPR}
\def\confYear{2026}

\title{Mining Instance-Centric Vision–Language Contexts for Human–Object Interaction Detection}
\author{
Soo Won Seo$^{1}$\thanks{Equal contributions.} \quad
KyungChae Lee$^{1}$\footnotemark[1] \quad
Hyungchan Cho$^{1}$ \quad
Taein Son$^{2}$ \quad\\
Nam Ik Cho$^{1}$ \quad
Jun Won Choi$^{1}$\thanks{Corresponding author.}\\
$^{1}$Seoul National University \quad
$^{2}$Hanyang University\\
{\small\texttt{
swseo@adr.snu.ac.kr \quad
\{kyungchae.lee, hyungchan1229, nicho, junwchoi\}@snu.ac.kr
}}\\
{\small\texttt{tison@spa.hanyang.ac.kr}}
}
\begin{document}
\maketitle
\begin{abstract}
Human–Object Interaction (HOI) detection aims to localize human–object pairs and classify their interactions from a single image, a task that demands strong visual understanding and nuanced contextual reasoning. 
Recent approaches have leveraged Vision–Language Models (VLMs) to introduce semantic priors, significantly improving HOI detection performance. 
However, existing methods often fail to fully capitalize on the diverse contextual cues distributed across the entire scene. 
To overcome these limitations, we propose the Instance-centric Context Mining Network (InCoM-Net)—a novel framework that effectively integrates rich semantic knowledge extracted from VLMs with instance-specific features produced by an object detector. 
This design enables deeper interaction reasoning by modeling relationships not only within each detected instance but also across instances and their surrounding scene context. 
InCoM-Net comprises two core components:
Instance-centric Context Refinement (ICR), which separately extracts intra-instance, inter-instance, and global contextual cues from VLM-derived features, and Progressive Context Aggregation (ProCA), which iteratively fuses these multi-context features with instance-level detector features to support high-level HOI reasoning. Extensive experiments on the HICO-DET and V-COCO benchmarks show that InCoM-Net achieves state-of-the-art performance, surpassing previous HOI detection methods. Code is available at \href{https://github.com/nowuss/InCoM-Net}{\texttt{https://github.com/nowuss/InCoM-Net}}.
\end{abstract}

\section{Introduction}
Human-Object Interaction (HOI) detection is a fundamental vision task that aims to localize all human and object instances in an image and classify the interaction types between them (e.g., holding, riding, pushing). Unlike conventional object detection, HOI detection requires  understanding of spatial configurations, object affordances, and contextual semantics to reason about the plausible relationships between humans and objects.
Recent Transformer-based architectures~\cite{kim2021hotr, zhang2022exploring, zhang2023exploring, cao2023re} have been proposed to advance HOI detection. However, despite these advancements, accurately capturing subtle, latent, and context-dependent interactions remains a considerable challenge.

The emergence of Vision-Language Models (VLMs) such as CLIP~\cite{radford2021learning}, BLIP~\cite{li2022blip}, and BLIP2~\cite{li2023blip} has enabled powerful joint understanding of visual and textual modalities. Trained on large-scale image-text pairs, VLMs have shown strong generalization capabilities in zero-shot image classification, captioning, and visual question answering. Owing to their ability to produce semantically rich visual representations aligned with language, recent studies~\cite{liao2022gen, mao2023clip4hoi, lei2024exploring} have applied VLMs to HOI detection, aiming to leverage high-level interaction priors.

\begin{figure}[!t]
\centering
\includegraphics[width=0.48\textwidth]{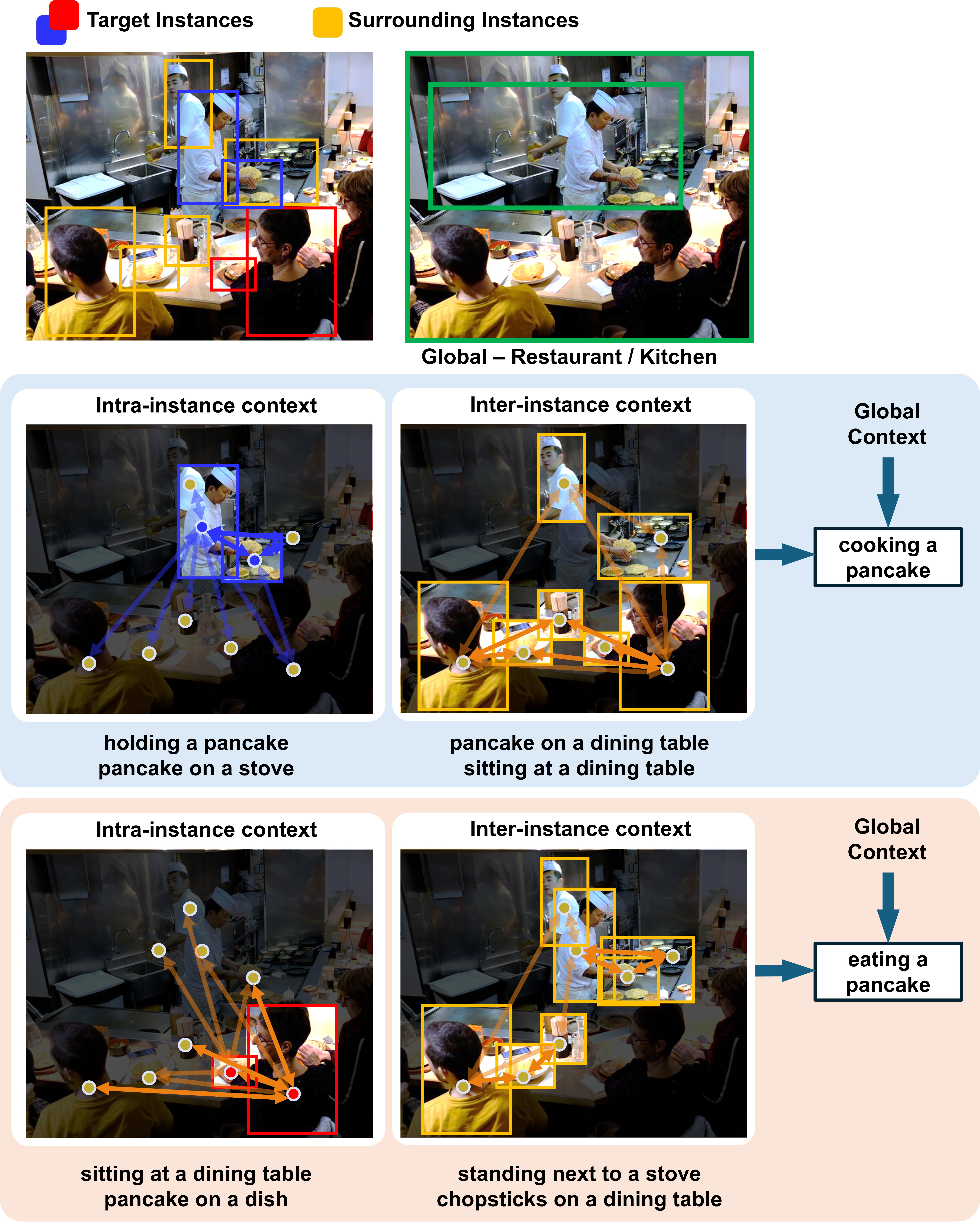}
\caption{
Illustration of three levels of contextual information.
For each instance, contextual information can be distinguished into intra-instance, inter-instance, and global contexts, each providing complementary cues for interpreting human–object interactions.
}
\label{fig:fig1_teas_real.pdf}
\end{figure}

Several studies have explored different strategies for utilizing features from VLMs in HOI detection. HOICLIP~\cite{ning2023hoiclip} introduced an interaction decoder that attends to VLM visual features to incorporate contextual cues. UniHOI~\cite{cao2023detecting} adopted a similar decoder structure and further improved generalization ability by using LLM-generated text as augmented textual guidance.
While these approaches focused on feature-level integration, several other methods aimed to exploit VLM features at the instance-level. ADA-CM~\cite{lei2023efficient} enhanced CLIP by injecting detection signals through lightweight adapters and applying instance-level pooling over adapted visual features. BCOM~\cite{wang2024bilateral} employed a dual-branch architecture that separately encodes RoI-aligned features from the object detector and the VLM. 

Despite recent progress, key challenges remain in HOI understanding. Effectively mining contextual information from VLM features is crucial for capturing human–object relationships. Various types of contextual cues can contribute to HOI reasoning, each playing a distinct yet complementary role.
As illustrated in Fig.~\ref{fig:fig1_teas_real.pdf}, humans rely on diverse contexts such as 1) visual cues from the target instance, 2) its relationships with surrounding instances, and 3) the broader surrounding scene context. Therefore, it is essential to effectively extract and integrate these multi-context features to enhance the model’s ability to infer complex human–object relationships.

\begin{figure*}[!t]
\centering
\includegraphics[width=0.97\textwidth]{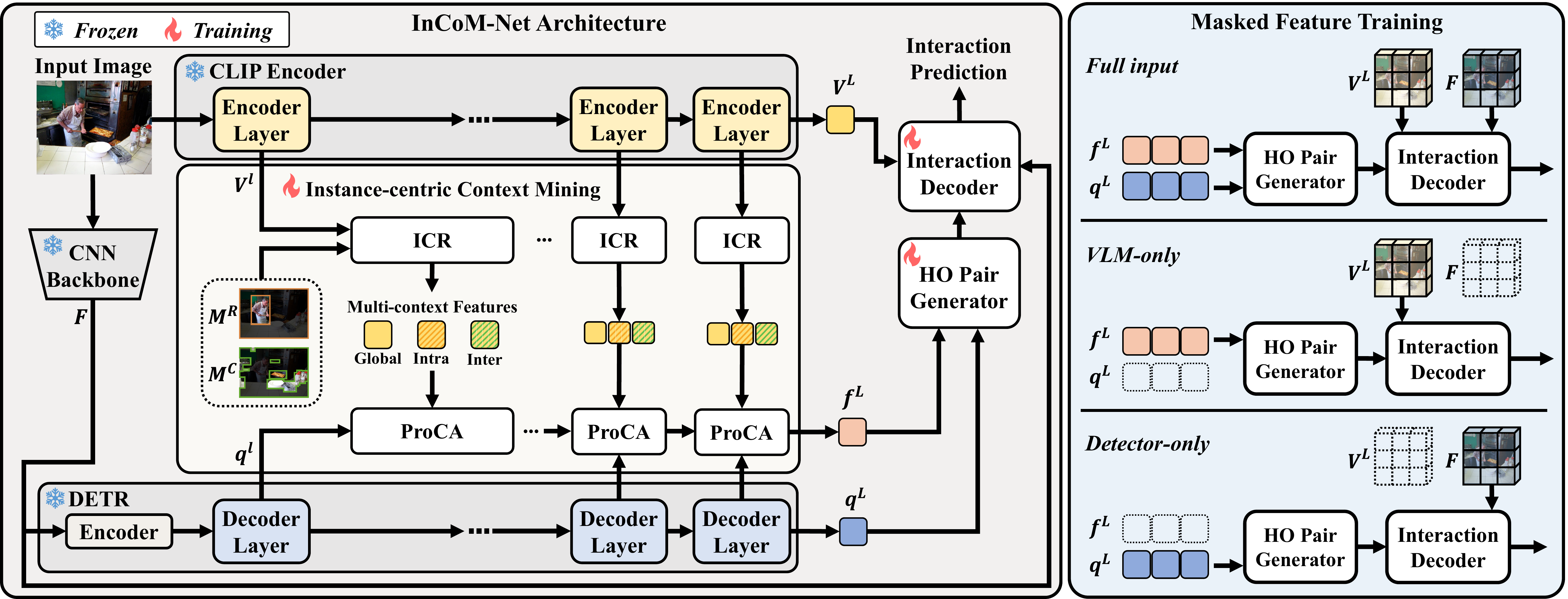}
\caption{Overall framework of InCoM-Net. Left: overview of Instance-centric Context Mining that integrates multi-context VLM features with instance-level features. Right: masked feature training (MFT), balancing the utilization of heterogeneous feature sources via masking.}

\label{fig:fig_overall}
\end{figure*}

Several methods have been proposed to utilize the context features for HOI detection. 
Most approaches~\cite{ning2023hoiclip,cao2023detecting,mao2023clip4hoi} have typically incorporated VLM representations in the final interaction reasoning phase as scene-level contextual cues.
Recent studies~\cite{lei2023efficient,lei2024exploring,kim2025locality,lei2025hola} inject detected instance information into VLM encoders to better align instance features with scene semantics.
However, these methods apply contextual information uniformly across all instances, failing to capture nuanced semantic information specific to each instance.
This limitation hinders their capability to accurately interpret complex human-object interactions.

To address these limitations, we propose the Instance-centric Context Mining Network (InCoM-Net), a novel framework for HOI detection. InCoM-Net constructs multi-context features for each instance using a VLM backbone and integrates them with instance-level features via Transformer-based attention. By modeling features across diverse contextual scopes, the network can be more effectively guided in both what and where to attend, enabling more robust and context-aware HOI understanding.

The main ideas of InCoM-Net are summarized as follows.
First, we introduce instance-centric multi-context feature generation through the Instance-centric Context Refinement (ICR) module. For each instance, three levels of contextual information—(1) intra-instance, (2) inter-instance, and (3) global contexts—are derived from VLM representations by applying scene-adaptive masks to the feature maps. These scene-adaptive masks enable the generation of context-specific features for each instance-level query. The intra-instance context captures visual cues within the target instance, the inter-instance context models interactions among multiple instances, and the global context provides complementary scene-level information to support comprehensive HOI reasoning. These multi-context features are encoded separately to preserve diverse semantic information and construct adaptive, multi-level contextual representations tailored to each instance.

Second, the multi-context features are aggregated into the instance-level features from the object detector through the Progressive Context Aggregation (ProCA) module. 
ProCA employs multiple attention layers to iteratively gather and refine relevant contextual information. This progressive integration enhances the alignment between each instance’s appearance and its surrounding context.

Third, we enable balanced utilization of both VLM multi-context features and detector spatial features through masked feature training (MFT). We observe that, when combining these heterogeneous feature sources for HOI prediction, the model tends to over-rely on one source, which limits overall performance. During training, MFT mitigates this imbalance by selectively masking either the VLM multi-context features or the detector spatial features. This strategy encourages the model to effectively leverage diverse contextual cues under varying input conditions.

We evaluate InCoM-Net on two widely used HOI benchmark datasets: HICO-DET~\cite{chao2018learning} and V-COCO~\cite{gupta2015visual}. 
The proposed InCoM-Net achieves state-of-the-art performance on both benchmarks. 
On HICO-DET, InCoM-Net achieves a {\bf +1.03} mAP improvement over the current best method, NMSR~\cite{yang2025no}. On V-COCO, our model also surpasses NMSR by a remarkable margin of \textbf{+3.8} mAP. Moreover, our method demonstrates superior generalization and robustness under the zero-shot setting of HICO-DET.

Our contributions can be summarized as follows:
\begin{itemize}
\itemsep 1em

\item We present InCoM-Net, a novel HOI detection framework that effectively aggregates multi-contextual cues to enable fine-grained interaction reasoning.

\item We propose a novel architecture that generates multi-context features from VLM representations and integrates them in an iterative manner. Specifically, we introduce a masking strategy to derive (1) intra-instance, (2) inter-instance, and (3) global contextual features. Both the masking and feature generation processes are conducted independently for each instance, enabling instance-specific context modeling.

\item We also introduce a masked feature training strategy that balances the utilization of heterogeneous feature sources, allowing the model to robustly exploit complementary information from both VLM and detector spatial features.

\item InCoM-Net achieves state-of-the-art performance on HICO-DET and V-COCO, demonstrating its effectiveness in capturing rich contextual cues for HOI detection.

\end{itemize}

\section{Related Work}

\subsection{Human-Object Interaction Detection}
HOI detection methods are broadly categorized into one-stage and two-stage methods.
One-stage methods~\cite{chen2021reformulating, zou2021end, kim2023relational, liao2022gen, jia2025contexthoi, zhang2022exploring} jointly detect objects and infer interactions within a unified framework.
Early approaches~\cite{liao2020ppdm, kim2020uniondet} adopt spatial anchors to associate human-object pairs in a single pass.
QPIC~\cite{tamura2021qpic} and HOTR~\cite{kim2021hotr} further advance this line of research by adopting the encoder-decoder architecture of DETR~\cite{carion2020end} for query-based reasoning.

In contrast, two-stage methods~\cite{qi2018learning, zhang2022efficient, lei2025hola, zhang2023exploring} first detect human and object instances, and then classify interactions between candidate pairs.
To capture broader context, graph-based approaches~\cite{ulutan2020vsgnet, gao2020drg, zhang2021spatially, hong2025learning} model relational structures by propagating information between human and object nodes using message-passing mechanisms.
More recent methods~\cite{zhang2023exploring, cao2023re} adopt Transformer architectures to improve context modeling, enabling more flexible encoding of global dependencies.

\subsection{VLM Integration for HOI Detection}
The emergence of VLMs pretrained on large-scale image–text pairs has led to growing interest in applying them to HOI detection. Recent studies~\cite{lei2023efficient, mao2023clip4hoi, ning2023hoiclip, cao2023detecting, lei2024ez, wang2024bilateral, jia2025contexthoi, geng2025horp,jia2025orchestrating} have explored diverse strategies for incorporating VLMs into HOI detection frameworks. Broadly, these approaches can be grouped into two categories.
The first line of work leverages scene-level VLM features as global semantic priors to guide interaction reasoning~\cite{ning2023hoiclip, mao2023clip4hoi, cao2023detecting, lei2024ez}.
The second line of research focuses on adapting VLM features to object instances~\cite{lei2023efficient, wang2024bilateral, lei2025hola,kim2025locality}. 
Although these methods improve locality awareness, their reliance on rigid RoI-based operations confines adaptation to object-bound regions and limits broader interaction context.

\begin{figure}[!t]
\centering
\includegraphics[width=0.46\textwidth]{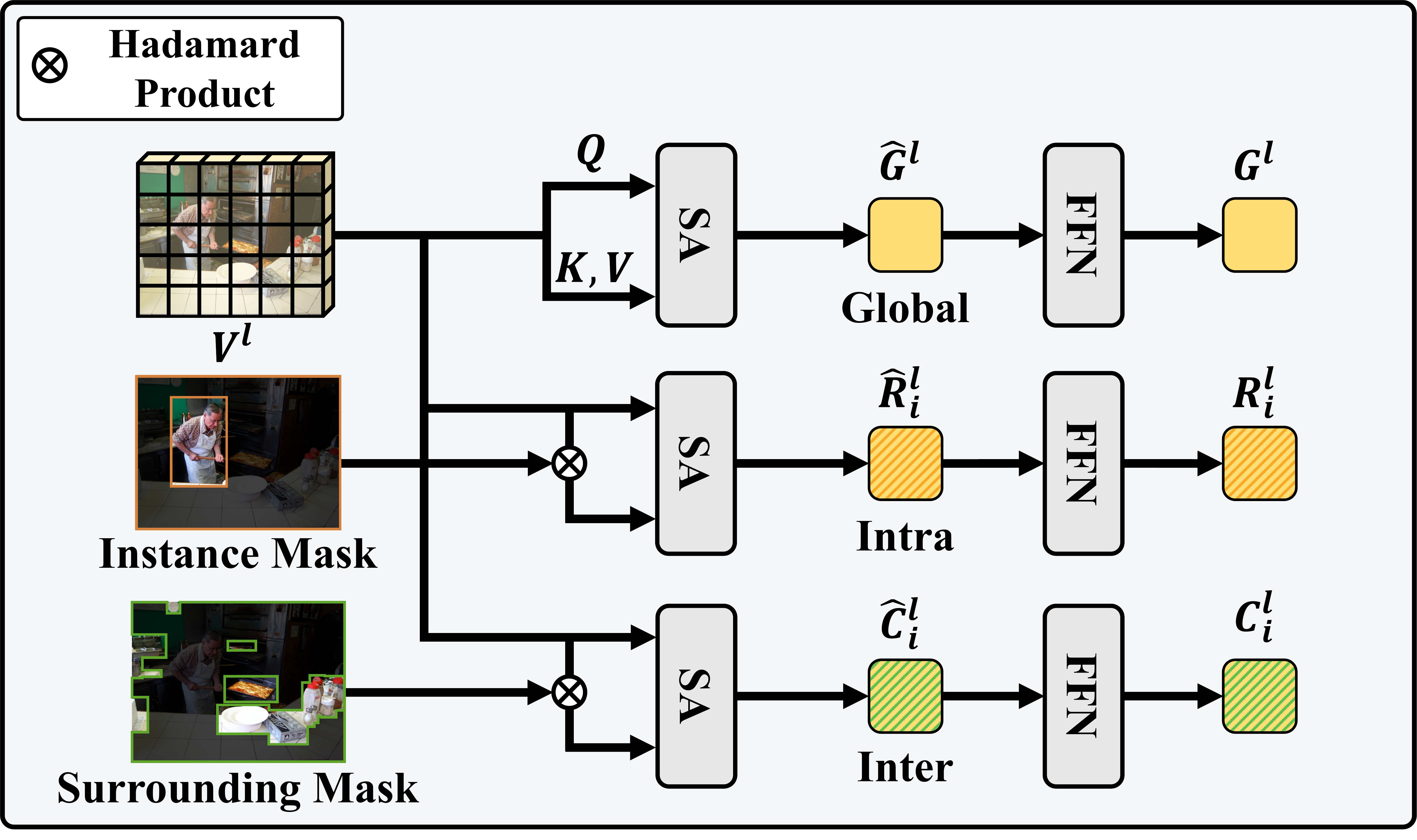}
\caption{Structure of ICR. The illustration shows the process of generating the multi-context features for the $i$-th instance.}
\label{fig:fig_icr}
\end{figure}

\section{Method}
\subsection{Overview}

The overall framework of InCoM-Net is illustrated in Fig.~\ref{fig:fig_overall}. Two independent backbone features are generated via two parallel pipelines: a DETR-based detector and a CLIP visual encoder.

First, a spatial feature map $F$ is extracted using a CNN backbone~\cite{he2016deep}, which is subsequently processed by DETR~\cite{carion2020end} to detect individual human and object instances. From the final $L$ decoder layers of DETR, we obtain instance-wise detector features denoted as $\{q^l\}_{l=1}^{L}$. In parallel, CLIP~\cite{radford2021learning} is used to extract VLM features. Specifically, we use the top $L$ layer features from the CLIP visual encoder, denoted as $\{V^l\}_{l=1}^{L}$, to serve as the VLM features.

InCoM-Net adopts Instance-centric Context Mining to model human-object-context interactions. This module is composed of two submodules: ICR and ProCA. 
These two modules operate over $L$ layers to produce context-aggregated features.
The ICR module first applies masked self-attention to the VLM features ${V^l}$, where $l$ denotes the layer index. 
The attention process is spatially guided by instance and surrounding masks from the detector.
Using these masks, ICR extracts intra-instance, inter-instance, and global context features for each instance from the VLM representations. 
The ProCA module then performs instance-centric cross-attention, using the detector query features $q^l$ as queries and the multi-context features from ICR as keys and values. This process is iteratively applied across all $L$ layers, resulting in context-aggregated features $f^l$.

Finally, the Human-Object (HO) pair generator constructs HO features for all possible human-object pairs by combining the instance-wise features $q^L$ from DETR with the context-aggregated features $f^L$. These HO features are then fed into the interaction decoder as queries for pairwise interaction modeling. The decoder applies cross-attention separately to CNN backbone features $F$ and VLM features $V^L$, and outputs interaction classification results.

To train InCoM-Net, we adopt a masked feature training strategy in which either detector-based or VLM-based features are masked. During training, three input configurations are applied equally: full input, detector-only, or VLM-only. This strategy encourages robust and balanced utilization of both feature sources.

\begin{figure}[!t]
\centering
\includegraphics[width=0.46\textwidth]{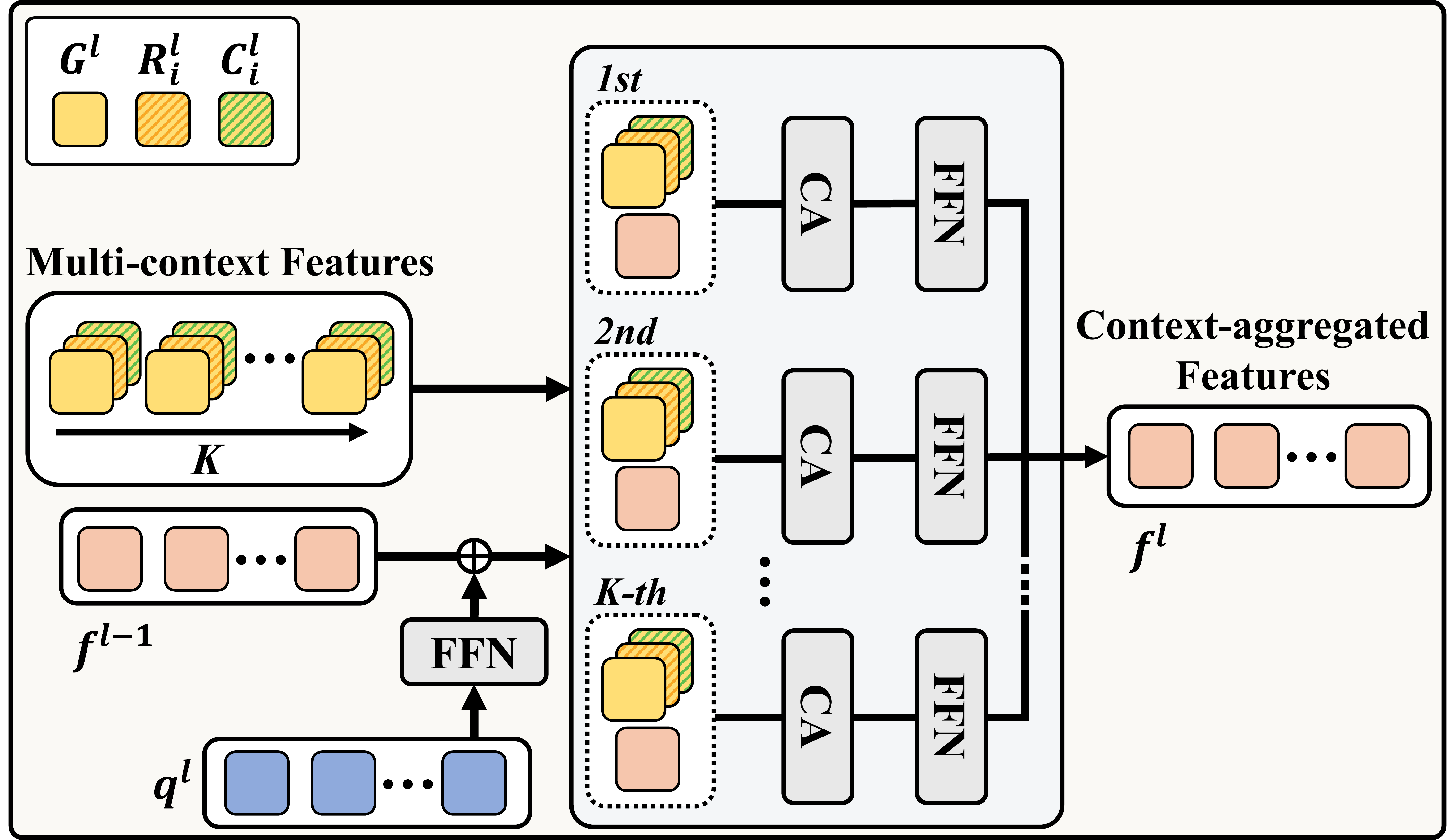}
\caption{Structure of ProCA.}
\label{fig:fig_proca}
\end{figure}

\subsection{Instance-centric Context Mining}

\paragraph{Instance-centric Context Refinement (ICR).}
The structure of ICR is illustrated in Fig.~\ref{fig:fig_icr}. ICR refines VLM context features to generate intra-instance, inter-instance, and global context features for each instance. Specifically, it applies self-attention over the VLM features ${V}^l \in \mathbb{R}^{N \times D_v}$, guided by instance masks $M^R \in \mathbb{R}^{K \times N}$ and surrounding masks $M^C \in \mathbb{R}^{K \times N}$ obtained from the detector, where $K$ and $N$ denote the number of detected instances and the sequence length of the VLM features, respectively. The instance mask $M^R_i$ is a binary mask indicating the region of the $i$-th instance. The surrounding mask $M^C_i$ is defined as the union of all other instance masks, excluding $M^R_i$.

Using the masks $M^R_i$ and $M^C_i$, the global context features $\hat{G}^l$, the intra-instance context features $\hat{R}^l_i$ and the inter-instance context features $\hat{C}^l_i$ are obtained as follows
\begin{align}
\hat{G}^l &= \text{Self-Attention}(V^l,\mathbf{1}), \\
\hat{R}_i^l &= \text{Self-Attention}(V^l, M^R_i), \\
\hat{C}_i^l &= \text{Self-Attention}(V^l, M^C_i),
\end{align}
where $\text{Self-Attention}(V, M)$ denotes the masked self-attention operation similar to~\citet{hajimiri2025pay}, with $V$ as the input feature and $M$ as the attention mask. The final multi-context representations are obtained by passing each attended feature through a feed-forward network (FFN)
\begin{align}
G^l &= \text{FFN}(\hat{G}^l), \\
R_i^l &= \text{FFN}(\hat{R}_i^l), \\
C_i^l &= \text{FFN}(\hat{C}_i^l).
\end{align}

\paragraph{Progressive Context Aggregation (ProCA).}
The structure of ProCA is shown in Fig.~\ref{fig:fig_proca}. 
ProCA forms context-aggregated features by leveraging cross-attention, where the detector queries attend to the multi-context features produced by ICR.
For the $i$-th instance, the detector query features $q_i^l$ are first combined with the context-aggregated features $f_i^{l-1}$ from the previous layer
\begin{align}
    \hat{f}_i^l &= {f}_i^{l-1} + \text{FFN}\left(q_i^l\right),
\end{align}
where $f^0_i$ is initialized as zero embedding.

Then, each combined instance feature $\hat{f}^l_i$ interacts with its global, intra-instance and inter-instance context features $G^l$, $R^l_i$, and $C^l_i$ to capture complementary semantics across multiple contextual levels. Specifically, cross-attention is applied to each context feature separately
\begin{align}
f_{i,G}^{l} &= \text{Cross-Attention}\left(\hat{f}_i^{l}, G^l, G^l\right),\\
f_{i,R}^{l} &= \text{Cross-Attention}\left(\hat{f}_i^{l}, R_i^l, R_i^l\right), \\
f_{i,C}^{l} &= \text{Cross-Attention}\left(\hat{f}_i^{l}, C_i^l, C_i^l\right),
\end{align}
where $\text{Cross-Attention}(Q, K, V)$ denotes the standard cross-attention operation with query $Q$, key $K$, and value $V$. 
The individual outputs are concatenated and passed through an FFN to produce the context-aggregated feature
\begin{align}
f_i^l = \text{FFN}\!\left([\,f_{i,G}^l \,\Vert\, f_{i,R}^l \,\Vert\, f_{i,C}^l\,]\right),
\end{align}
where $\vert\vert$ denotes channel-wise concatenation.
This process is repeated for $L$ layers, enabling ProCA to progressively integrate multi-level semantic features from the VLM and produce the final context-aggregated features $f_i^L$.

\subsection{HO Pair Generation and Interaction Reasoning}
\label{sec:HO_pair_gen}
\paragraph{HO Pair Generator.}
We construct HO pair features $s$ by fusing detector query features $q^L$ and context-aggregated features $f^L$.
For each human-object pair $(h, o)$, we combine corresponding detector query features $(q_h^L, q_o^L)$ and context-aggregated features $(f^L_h, f^L_o)$

\begin{align}
    {s}  &= \text{LN}\left(\text{Linear}\left({q}^L_{h}\mathbin{\Vert} {q}^L_{o}\right)\right) \notag \\
     &\quad + \text{LN}\left(\text{Linear}\left(f^{L}_{h}\mathbin{\Vert} f^L_{o}\right)\right),
\end{align}
where LN denotes layer normalization.
The resulting pair features $s$ are then passed through a shallow Transformer encoder to obtain the final HO pair feature $\hat{s}$.

\paragraph{Interaction Decoder.}
The interaction decoder enhances the HO pair features $\hat{s}$ for HOI reasoning. 
The decoder queries are initialized to $z^0 = \hat{s}$ and updated through $L_z$ layers. 
At the $l$-th layer, $z^l$ is processed by a self-attention block and two parallel cross-attention blocks, which attend to the CNN backbone features $F$ and the VLM features $V^L$, respectively
\begin{align}
    \hat{z}^l  =& \text{Self-Attention}({z^l}), \\
    z^{l+1}  = &\text{FFN}(\text{Cross-Attention}(\hat{z}^l, F, F)  \nonumber \\
    &+ \text{Cross-Attention}(\hat{z}^l, V^L, V^L)),
\end{align}
The final decoder outputs are passed to the interaction classifier to obtain the final results.

\subsection{Masked Feature Training (MFT)}
\label{sec:mft}
We train InCoM-Net using MFT, as shown in Fig.~\ref{fig:fig_overall} (right). Specifically, we construct three masked input configurations: (1) \textit{full input} $x_f=\{f^L,q^L, V^L, F\}$, (2) \textit{detector-only} $x_d=\{q^L, F\}$, and (3) \textit{VLM-only} $x_v=\{f^L, V^L\}$. Masking is applied to both the instance-wise features and the interaction decoder inputs. When masked, instance features are zeroed out, and the corresponding cross-attention blocks are deactivated to yield zero outputs.

During training, each configuration is individually processed by the sub-module $\Phi_{\theta}(\cdot)$, which consists of the HO pair generator, the interaction decoder, and the interaction classifier. The final training objective is defined as follows

\begin{align}
  \mathcal{L} = \sum_{x \in \mathcal{X}}\mathcal{L}_f\left(y,\, \Phi_{\theta}(x)\right),
\end{align}
where $\mathcal{X} = \{x_f, x_d, x_v\}$, $\mathcal{L}_f$ denotes the focal loss, and $y$ is the ground-truth interaction label.

\begin{table*}[t]
    \caption{Performance comparison under the regular setting on the HICO-DET and V-COCO datasets. R50 denotes ResNet-50.}
    \centering
    \begin{adjustbox}{width=0.95\textwidth}
    \begin{tabular}{llcccccccc}
        \toprule 
        & & \multicolumn{6}{c}{\textbf{HICO-DET}} 
        & \multicolumn{2}{c}{\textbf{V-COCO}} \\
        & & \multicolumn{3}{c}{\textbf{Default Setting}} 
          & \multicolumn{3}{c}{\textbf{Known-Object Setting}} \\
        \cmidrule(lr){3-5} \cmidrule(lr){6-8}
        {\textbf{Method}} & {\textbf{Backbone}} & {\textbf{Full}} & {\textbf{Rare}} & {\textbf{Non-rare}} & {\textbf{Full}} & {\textbf{Rare}} & {\textbf{Non-rare}} & \textbf{$\mathbf{AP^{\textbf{S1}}_{\textbf{role}}}$} & \textbf{$\mathbf{AP^{\textbf{S2}}_{\textbf{role}}}$}\\
        \midrule
        QPIC~\cite{tamura2021qpic} \textsubscript{\textcolor{gray}{[CVPR21]}} & R50 & 29.07 & 21.85 & 31.23 & 31.68 & 24.14 & 33.93 & 58.8 & 61.0 \\
        UPT~\cite{zhang2022efficient} \textsubscript{\textcolor{gray}{[CVPR22]}} & R50 & 31.66 & 25.94 & 33.36 & 35.05 & 29.27 & 36.77 & 59.0 & 64.5 \\
        STIP~\cite{zhang2022exploring} \textsubscript{\textcolor{gray}{[CVPR22]}} & R50 & 32.22 & 28.15 & 33.43 & 35.29 & 31.43 & 36.45 & 66.0 & 70.7 \\
        PViC~\cite{zhang2023exploring} \textsubscript{\textcolor{gray}{[ICCV23]}} & R50 & 34.69 & 32.14 & 35.45 & 38.14 & 35.38 & 38.97 & 62.8 & 67.8 \\
        RmLR~\cite{cao2023re} \textsubscript{\textcolor{gray}{[ICCV23]}} & R50 & 36.93 & 29.03 & 39.29 & 38.29 & 31.41 & 40.34 & 63.8 & 69.8 \\
        
        \midrule
        \rowcolor{Gray}
        \multicolumn{10}{l}{\textbf{VLM-based Methods with ViT-B}} \\
        GEN-VLKT~\cite{liao2022gen} \textsubscript{\textcolor{gray}{[CVPR22]}} & R50+CLIP & 33.75 & 29.25 & 35.10 & 36.78 & 32.75 & 37.99 & 62.4 & 64.5 \\
        HOICLIP~\cite{ning2023hoiclip} \textsubscript{\textcolor{gray}{[CVPR23]}} & R50+CLIP & 34.69 & 31.12 & 35.74 & 37.61 & 34.47 & 38.54 & 63.5 & 64.8 \\
        CLIP4HOI~\cite{mao2023clip4hoi} \textsubscript{\textcolor{gray}{[NeurIPS23]}} & R50+CLIP & 35.33 & 33.95 & 35.74 & 37.19 & 35.27 & 37.77 & -- & 66.3 \\
        HOLa~\cite{lei2025hola} \textsubscript{\textcolor{gray}{[ICCV25]}} & R50+CLIP & 35.41 & 34.35 & 35.73 & -- & -- & -- & -- & -- \\
        LAIN~\cite{kim2025locality} \textsubscript{\textcolor{gray}{[CVPR25]}} & R50+CLIP & 36.02 & 35.70 & 36.11 & -- & -- & -- & -- & 65.1 \\
        GroupHOI~\cite{hong2025learning}  \textsubscript{\textcolor{gray}{[NeurIPS25]}} & R50+CLIP & 36.70 & 34.86 & 37.26 & 39.42 & 37.78 & 39.91 & 65.0 & 66.0 \\
        SCTC~\cite{jiang2024exploring} \textsubscript{\textcolor{gray}{[AAAI24]}} & R50+CLIP & 37.92 & 34.78 & 38.86 & -- & -- & -- & \underline{67.1} & \underline{71.7} \\
        HORP~\cite{geng2025horp} \textsubscript{\textcolor{gray}{[CVPR25]}} & R50+CLIP & \underline{38.61} & \underline{36.14} & \underline{39.34} & \underline{40.98} & \underline{38.25} & \underline{41.79} & 65.6 & 68.3 \\
        
        \midrule
        \textbf{InCoM-Net} &  R50+CLIP & \textbf{39.53} & \textbf{38.87} & \textbf{39.73} & \textbf{42.24} & \textbf{40.50} & \textbf{42.76} & \textbf{72.2} & \textbf{74.2} \\
        \midrule
        \rowcolor{Gray}
        \multicolumn{10}{l}{\textbf{VLM-based Methods with ViT-L}} \\
        CMMP~\cite{lei2024exploring} \textsubscript{\textcolor{gray}{[ECCV23]}} & R50+CLIP & 38.14 & 37.75 & 38.25 & -- & -- & -- & -- & 64.0 \\
        ADA-CM~\cite{lei2023efficient} \textsubscript{\textcolor{gray}{[ICCV23]}} & R50+CLIP & 38.40 & 37.52 & 38.66 & -- & -- & -- & 58.6 & 64.0 \\
        HOLa~\cite{lei2025hola} \textsubscript{\textcolor{gray}{[ICCV25]}} & R50+CLIP & 39.05 & 38.66 & 39.17 & -- & -- & -- & 60.3 & 66.0 \\
        VDRP~\cite{yang2025visual} \textsubscript{\textcolor{gray}{[NeurIPS25]}} & R50+CLIP & 39.07 & 39.08 & 39.06 & -- & -- & -- & 60.6 & 66.2 \\
        BCOM~\cite{wang2024bilateral} \textsubscript{\textcolor{gray}{[CVPR24]}} & R50+CLIP & 39.34 & 39.90 & 39.17 & {42.24} & {42.86} & 42.05 & 65.8 & {69.9} \\
        GroupHOI~\cite{hong2025learning} \textsubscript{\textcolor{gray}{[NeurIPS25]}} & R50+CLIP & 39.46 & 37.10 & 40.16 & 41.58 & 39.42 & {42.40} & 66.4 & 67.3 \\
        UniHOI~\cite{cao2023detecting} \textsubscript{\textcolor{gray}{[NeurIPS23]}} & R50+BLIP2 & 40.06 & 39.91 & 40.11 & 42.20 & 42.60 & 42.08 & 65.6 & 68.3 \\
        ContextHOI~\cite{jia2025contexthoi} \textsubscript{\textcolor{gray}{[AAAI25]}} & R50+CLIP & 41.82 & 43.91 & 41.19 & -- & -- & -- & 66.1 & -- \\
        SPLHOI~\cite{luo2025synergistic} \textsubscript{\textcolor{gray}{[TIP25]}}  & R50+BLIP2 & {42.06} & {42.53} & {41.92} & 44.62 & \underline{45.49} & 44.36 & {68.1} & \underline{72.9} \\
        InterProDa~\cite{jia2025orchestrating} \textsubscript{\textcolor{gray}{[AAAI25]}}  & R50+CLIP & {42.67} & \underline{45.21} & {41.92} & -- & -- & -- & {67.6} & -- \\
        NMSR~\cite{yang2025no} \textsubscript{\textcolor{gray}{[ICCV25]}}  & R50+CLIP & \underline{42.93} & {42.41} & \underline{43.11} & \underline{44.97} & 44.20 & \underline{45.23} & \underline{69.8} & 72.1 \\
        \midrule
        \textbf{InCoM-Net} &  R50+CLIP & \textbf{43.96} & \textbf{45.61} & \textbf{43.46} & \textbf{45.59} & \textbf{45.92} & \textbf{45.49} & \textbf{73.6} & \textbf{75.4} \\
    \bottomrule
    \end{tabular}
    \end{adjustbox}
    \label{tab:overall}
\end{table*}

\section{Experiments}
\subsection{Datasets and Evaluation Settings}
\paragraph{Datasets.} We evaluate InCoM-Net on two widely used benchmarks for HOI detection: HICO-DET~\cite{chao2018learning} and V-COCO~\cite{gupta2015visual}.
HICO-DET contains 47,776 images, including 38,118 training images and 9,658 test images, with annotations for 600 HOI categories defined by 117 verbs and 80 object classes. V-COCO, a subset of MS-COCO~\cite{lin2014microsoft}, consists of 10,346 images annotated with 26 common actions, each potentially associated with one or more object roles.

\paragraph{Regular Settings.} We follow standard evaluation protocols, reporting the mean Average Precision (mAP). 
A prediction is considered correct when both the human and object bounding boxes have an Intersection over Union (IoU) greater than 0.5 with the ground truth, and the interaction class is correctly predicted.
For HICO-DET, we report mAP on three standard splits: \textit{Full} (all 600 categories), \textit{Rare} (138 categories with fewer than 10 training instances), and \textit{Non-Rare} (the remaining 462 categories). For V-COCO, we report role-based mAP ($\text{AP}_{\text{role}}$), requiring correct prediction of both the action and its object roles with IoU $\geq$ 0.5, under both Scenario 1 and Scenario 2.

\paragraph{Zero-shot Settings.} 
In the zero-shot HOI detection setting, the 600 HOI categories in HICO-DET are split into 480 seen and 120 unseen categories. The model is trained only on the seen categories and evaluated on both seen and unseen categories at test time.
We report results under two settings: Rare First Unseen Combination (RF-UC) and Non-rare First Unseen Combination (NF-UC).
In RF-UC, the 120 unseen categories are sampled from rare interactions, whereas in NF-UC, they are selected from non-rare interactions. 
The evaluation procedure follows the standard HICO-DET protocol, except that performance is reported on the \textit{Unseen}, \textit{Seen}, and \textit{Full} splits.

\paragraph{Implementation Details.}
We adopted DETR~\cite{carion2020end} with ResNet-50~\cite{he2016deep} backbone, pre-trained on MS-COCO and fine-tuned on the target HOI dataset. For the CLIP encoder~\cite{radford2021learning}, we considered two visual backbones: ViT-B and ViT-L. Both DETR and CLIP were kept frozen during training. The number of interaction decoder layers and the value of $L$ were set to 2 and 3, respectively. The model was optimized using AdamW with a weight decay of $10^{-4}$. The initial learning rate was set to $10^{-4}$ and was reduced by a factor of 5 every 10 epochs. Training was performed for 30 epochs with a batch size of 16 on four NVIDIA GeForce RTX 3090 GPUs. Other implementation details are provided in the \textit{Supplementary Material}.

\begin{table}[t]
    \centering
    \caption{Zero-shot performance comparison on HICO-DET.}
    \setlength{\tabcolsep}{1mm} 
    \fontsize{9}{10}\selectfont
    \begin{tabular}{lcccccc}
        \toprule
          & \multicolumn{3}{c}{\textbf{RF-UC}} 
          & \multicolumn{3}{c}{\textbf{NF-UC}} \\
        \textbf{Method}                               
& \textbf{Unseen} & \textbf{Seen} & \textbf{Full} 
               & \textbf{Unseen} & \textbf{Seen} & \textbf{Full} \\
        \midrule
\rowcolor{Gray}
\multicolumn{7}{l}{\textbf{VLM-based Methods with ViT-B}} \\
        GEN-VLKT~\cite{liao2022gen}                     
& 21.36           & 32.91          & 30.56          
& 25.05           & 23.38          & 23.71          
\\
 HOICLIP~\cite{ning2023hoiclip}                      
& 25.53           & 34.85          & 32.99          
& 26.39           & 28.10          &27.75          
\\

 ADA-CM~\cite{lei2023efficient}                       
& 27.63           & 34.35          & 33.01          
& 32.41           & 31.13          &31.39          
\\
 CLIP4HOI~\cite{mao2023clip4hoi}                     
& 28.47           & \underline{35.48}          & 34.08          
& 31.44           & 28.26          &28.90          
\\
 CMMP~\cite{lei2024exploring}                         
& 29.45           & 32.87          & 32.18          
& 32.09           & 29.71          &30.18          
\\

        HOLa~\cite{lei2025hola}                         
& 30.61           & 35.08          & 34.19          
& 35.25           & 31.64          & 32.36          
\\
        VDRP~\cite{yang2025visual}
& 31.29& 34.41& 33.78
& \underline{36.45}& 31.60& 32.57
\\
 LAIN~\cite{kim2025locality}                         
& \underline{31.83}           & 35.06          & \underline{34.41}          
& 36.41           & \underline{32.44}          &\underline{33.23}          
\\
\midrule
 \textbf{InCoM-Net} & \textbf{32.65}& \textbf{38.69}& \textbf{37.48}& \textbf{38.68}& \textbf{35.00}& \textbf{35.74}\\
\midrule
\rowcolor{Gray}
\multicolumn{7}{l}{\textbf{VLM-based Methods with ViT-L}} \\

 BCOM~\cite{wang2024bilateral}             
& 28.52           & 35.04          & 33.74          
& 33.12           & 31.76          &32.03          
\\
 UniHOI~\cite{cao2023detecting}          
& 28.68           & 33.16          & 32.27          
& 28.45           & 32.63          &31.79          
\\
 CMMP~\cite{lei2024exploring}             
& 35.98           & 37.42          & 37.13          
& 33.52           & 35.53          & 35.13          
\\
 InterProDA~\cite{jia2025orchestrating}             
& 36.38           & \underline{40.88}          & \underline{39.58}          
& 33.64           & \underline{36.47}          & 35.50          
\\
 LAIN~\cite{kim2025locality}             
& 36.57           & 38.54          & 38.13          
& \underline{37.52}           & 35.90          &36.22          
\\
 VDRP~\cite{yang2025visual}
& \underline{36.72}           & 38.48          & 38.13
& 37.48           & 36.21          & \underline{36.46}
\\
\midrule

        \textbf{InCoM-Net}& \textbf{37.69}           & \textbf{41.46}          & \textbf{40.71}          & \textbf{39.45}            & \textbf{38.42}          & \textbf{38.62}          \\
    \bottomrule
    \end{tabular}
    \label{tab:zs_results2}
\end{table}

\subsection{Performance Comparison}
\paragraph{Regular HOI Detection.}

We compare InCoM-Net with existing HOI detection methods on two widely used benchmarks. As shown in Table~\ref{tab:overall}, InCoM-Net achieves state-of-the-art performance on both HICO-DET and V-COCO.

On HICO-DET, our InCoM-Net surpasses the previous best method HORP~\cite{geng2025horp} by 0.92 mAP with ViT-B, and outperforms NMSR~\cite{yang2025no} by 1.03 mAP using ViT-L.

On V-COCO, with ViT-B, InCoM-Net achieves 72.2 $\text{AP}^{\text{S1}}_{\text{role}}$ and 74.2 $\text{AP}^{\text{S2}}_{\text{role}}$, outperforming the previous best method SCTC~\cite{jiang2024exploring} by 5.1 $\text{AP}^{\text{S1}}_{\text{role}}$ and 2.5 $\text{AP}^{\text{S2}}_{\text{role}}$, respectively.
With ViT-L, the performance further improves to 73.6 $\text{AP}^{\text{S1}}_{\text{role}}$ and 75.4 $\text{AP}^{\text{S2}}_{\text{role}}$, surpassing NMSR~\cite{yang2025no} by {3.8} on $\text{AP}^{\text{S1}}_{\text{role}}$ and SPLHOI~\cite{luo2025synergistic} by {2.5} on $\text{AP}^{\text{S2}}_{\text{role}}$.

\paragraph{Zero-shot HOI Detection.}
\label{sec:zeroshot}
Table~\ref{tab:zs_results2} presents the zero-shot results on HICO-DET.
InCoM-Net consistently achieves state-of-the-art results under both RF-UC and NF-UC settings. With a ViT-B backbone, it surpasses the previous best method, LAIN~\cite{kim2025locality}, by 0.82 mAP on RF-UC Unseen, and outperforms VDRP~\cite{yang2025visual} by 2.23 mAP on NF-UC Unseen. When scaled to a ViT-L backbone, InCoM-Net further improves performance, exceeding VDRP~\cite{yang2025visual} by 0.97 mAP on RF-UC Unseen and LAIN~\cite{kim2025locality} by 1.93 mAP on NF-UC Unseen.
Additional quantitative results are provided in the \textit{Supplementary Material}.

\begin{figure*}[!t]
\centering
\includegraphics[width=0.94\textwidth]{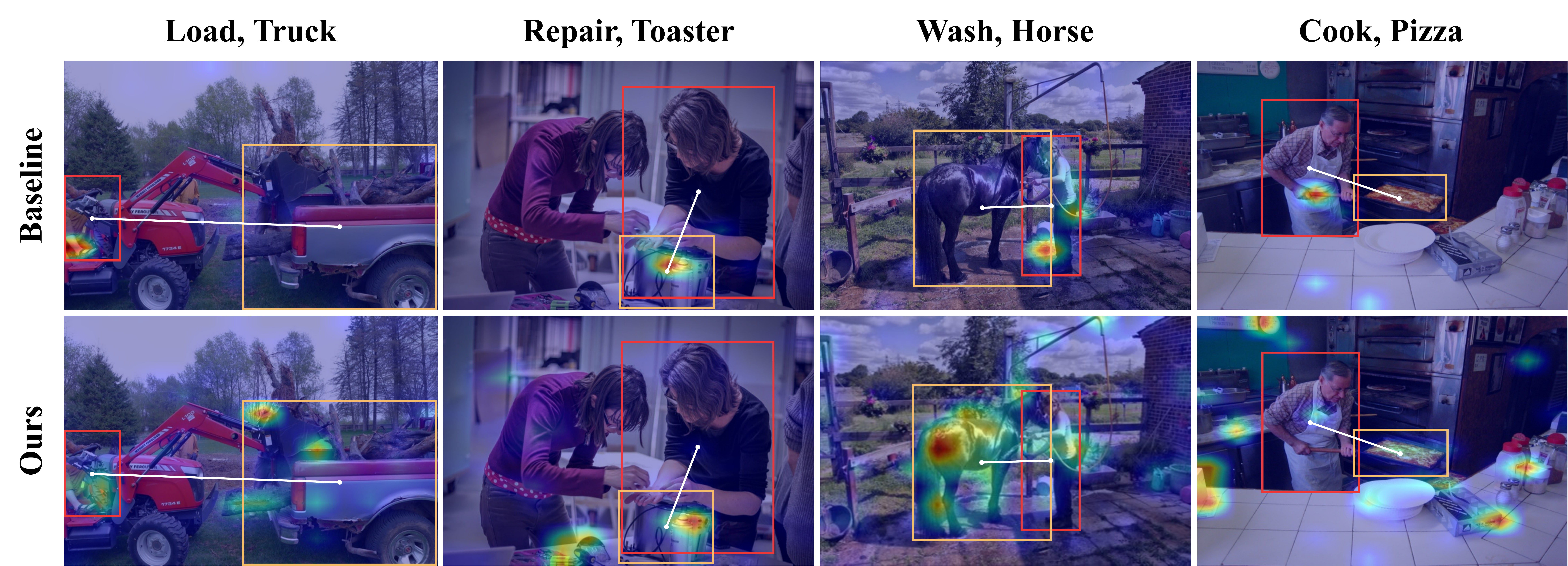}
\caption{Visualization of activation maps from the interaction decoder.}
\label{fig:qual}
\end{figure*}

\subsection{Ablation Study}
Our ablation studies were conducted on HICO-DET under the regular setting using InCoM-Net with ViT-B.

\paragraph{Contributions of Main Components.}
Table~\ref{tab:abl_components} presents the performance of the proposed method as each component is incrementally introduced.
The baseline model removes the ICR and ProCA from the InCoM-Net, using only the detector instance features as input to the HO Pair generator for interaction modeling.
Incorporating ICR improves performance by 1.25 mAP, demonstrating the benefit of multi-contextual representations.
Applying ProCA yields a further performance gain of 1.00 mAP, indicating its effectiveness in integrating diverse semantic cues from VLMs.
Finally, introducing MFT provides an additional 1.11 mAP improvement, underscoring the importance of balancing the utilization of heterogeneous feature sources.

\paragraph{Impact of Multi-context Modeling.}
Table~\ref{tab:abl_icr} reports the contribution of different context types in ICR. The baseline employs raw VLM features $(V)$ as a uniform context representation.
As each context type is added—global $(G)$, intra-instance $(R)$, and inter-instance $(C)$—performance consistently improves. In particular, incorporating intra-instance context yields a 0.54 mAP gain, with a notable improvement of 1.47 mAP on rare categories. Overall, the complete multi-context configuration achieves a cumulative 1.23 mAP improvement over the baseline, demonstrating that leveraging diverse context types effectively enriches scene representations for HOI detection.

\paragraph{Ablation Study of ProCA Layers.}
Table~\ref{tab:abl_proca} presents the results with different numbers of ProCA layers.
As the number of layers increases from $L$=1 to $L$=4, performance consistently improves, yielding a 1.06 mAP gain. This indicates that the ProCA effectively leverages semantic information from varying VLM layers. However, the gains become marginal beyond $L$=3; therefore, we adopt $L$=3 as the default configuration.

\begin{table}[t]
    \centering
    \caption{Ablation study of main components of InCoM-Net.}
    \fontsize{9}{10}\selectfont
    \begin{tabular}{L{1.7cm}C{1.2cm}C{1.2cm}C{1.2cm}}
        \toprule
        {Method} & {Full} & {Rare} & {Non-rare} \\
        \midrule
        Baseline & 36.17 & 33.11 & 37.08 \\
        + ICR     & 37.42  & 34.47 & 38.30    \\
        + ProCA   & 38.42 &  36.80 &  38.90 \\
        + MFT     & \textbf{39.53} & \textbf{38.87} & \textbf{39.73}\\
        \bottomrule
    \end{tabular}
    \label{tab:abl_components}
\end{table}

\begin{table}[t]
    \centering
    \caption{Effect of different context types in ICR.}

    \fontsize{9}{10}\selectfont
    \begin{tabular}{L{1.7cm}C{1.2cm}C{1.2cm}C{1.2cm}}
        \toprule
        {Context} & {Full} & {Rare} & {Non-rare} \\
        \midrule
        Baseline $(V)$               & 38.30 & 37.31  & 38.60  \\
        $G$               & 38.65  & 36.76 & 39.21  \\
        $G$, $R$           & 39.19  & 38.78 & 39.32  \\
        $G$, ${R}$, $C$       & \textbf{39.53} & \textbf{38.87} & \textbf{39.73}\\
        \bottomrule
    \end{tabular}

    \label{tab:abl_icr}
\end{table}

\begin{table}[t!]
    \centering
    \caption{Ablation study on the number of ProCA layers.}
    \begin{adjustbox}{width=0.85\linewidth}
    \fontsize{9}{10}\selectfont
    \begin{tabular}{C{1.4cm}C{1.3cm}C{1.3cm}C{1.3cm}}
        \toprule
        Layers & Full & Rare & Non-rare \\
        \midrule
        1    & 37.42 & 34.47 & 38.30 \\
        2    & 37.99 & 36.21 & 38.53 \\
        3    & 38.42 & 36.80 & 38.90\\
        4    & \textbf{38.48} & \textbf{37.06} & \textbf{38.92} \\
        \bottomrule
    \end{tabular}
    \label{tab:abl_proca}
    \end{adjustbox}
\end{table}

\paragraph{Impact of Masked Feature Training.}
To verify that MFT mitigates the imbalance in feature utilization, we apply it to our model, the baseline, and HOICLIP~\cite{ning2023hoiclip}, a representative VLM-based HOI method.
We first examine whether HOI models tend to favor one feature source over the other. Specifically, we evaluate all models under three inference settings: \textit{Full} (both VLM and detector features), \textit{V-only} (VLM features only), and \textit{D-only} (detector features only), while training all models with both features types. 

As shown in Table~\ref{tab:abl_mft}, without MFT, all methods suffer a substantial performance drop in the \textit{D-Only} setting, indicating a strong bias toward VLM features and confirming the imbalance issue.
MFT effectively alleviates this problem. Across all methods, MFT significantly narrows the gap between \textit{V-only} and \textit{D-only} settings and consistently improves overall performance by over 1.1 mAP. These results demonstrate that MFT encourages more balanced utilization of VLM and detector spatial features, leading to more robust HOI reasoning under varying input conditions.

\begin{table}[t!]
\centering
    \caption{Impact of MFT. * denotes reproduced results using their open-source code.}
\begin{adjustbox}{width=0.9\linewidth}
\fontsize{9}{10}\selectfont
\begin{tabular}{lcccc}
\toprule
Method & MFT & Full & V-only & D-only \\
\midrule
\multirow{2}{*}{HOICLIP$^{*}$~\cite{ning2023hoiclip}}   &            & 34.00 & 30.63 & 23.98 \\
                                                        & \checkmark & 35.41 & 34.48 & 32.72 \\
\midrule
\multirow{2}{*}{Baseline}                               &            & 36.17 & 33.21 & 18.52 \\
                                                        & \checkmark & 37.30 & 36.02 &  33.66  \\
\midrule
\multirow{2}{*}{\textbf{InCoM-Net}}                         &              & 38.42 & 36.20 & 19.11 \\
                                                        & \checkmark   & \textbf{39.53} & \textbf{38.41} & \textbf{36.14} \\
\bottomrule
    \end{tabular}

    \label{tab:abl_mft}
    \end{adjustbox}
\end{table}

\begin{table}[t!]
    \centering
    \caption{Comparison with alternative VLM feature extraction strategies.}
    \begin{adjustbox}{width=0.92\linewidth}
    \fontsize{9}{10}\selectfont
    \begin{tabular}{L{1.6cm}C{1.3cm}C{1.3cm}C{1.3cm}}
        \toprule
        Strategy & Full & Rare & Non-rare \\
        \midrule
        RoI Align & 36.37  & 34.48 &  36.93\\
        Img Crop  & 37.34  & 35.40 &  37.92\\
        \textbf{InCoM-Net}     & \textbf{38.42}  & \textbf{36.80} &  \textbf{38.90}\\
        \bottomrule
    \end{tabular}
    \label{tab:abl_extraction}
    \end{adjustbox}
\end{table}

\paragraph{Effect of Instance-centric Context Mining.}
Table~\ref{tab:abl_extraction} evaluates the effectiveness of Instance-centric Context Mining approach by replacing ICR and ProCA with two alternative strategies: RoI Align and Image Crop. RoI Align pools instance features from the VLM features using bounding box coordinates, while Image Crop independently crops and encodes each instance.
Both approaches extract instance features without incorporating surrounding context.
In contrast, our method constructs instance representations with their surrounding context through ICR and ProCA, outperforming RoI Align and Image Crop by 2.05 and 1.08 mAP, respectively. The performance gap is more pronounced on rare categories, reaching 2.32 and 1.40 mAP, underscoring the importance of instance-centric context modeling.

\subsection{Qualitative Results}
Fig.~\ref{fig:qual} presents a qualitative comparison between the activation maps of InCoM-Net and the baseline model without ICR and ProCA.
The baseline tends to focus on limited or irrelevant regions for the interaction. 
In contrast, our method attends to key interaction regions as well as contextual cues essential for HOI reasoning.

\section{Conclusion}
In this work, we proposed InCoM-Net, a novel HOI detection framework that leverages Instance-centric Context Mining to more effectively utilize semantically rich VLM representations for interaction reasoning.
By extracting multi-context information from VLM features, our model constructs adaptive contextual representations for each instance and progressively integrates them through a dedicated attention mechanism.
This allows the model to capture rich contextual cues that are critical for accurate HOI prediction. In addition, our masked feature training encourages balanced utilization of heterogeneous feature sources, enhancing robustness under varying input conditions. 
Extensive experiments on HICO-DET and V-COCO demonstrate that InCoM-Net achieves state-of-the-art performance and strong generalization in zero-shot settings. Overall, this work underscores the value of instance-centric context modeling for VLM-based HOI detection, enabling more effective use of semantically rich VLM representations.

\clearpage
\clearpage
\section{Acknowledgments}
This work was supported by 1) the Ministry of Trade, Industry and Energy (MOTIE, Korea) (No.RS-2024-00443216, Development and PoC of On-Device AI Computing based AI Fusion Mobility Device), and 2) Institute of Information \& communications Technology Planning \& Evaluation (IITP) grant funded by the Korea government(MSIT) [NO.RS-2021-II211343, Artificial Intelligence Graduate School Program (Seoul National University)]

{
    \small
    \bibliographystyle{ieeenat_fullname}
    \bibliography{main}
}

\end{document}